\documentclass[conference]{IEEEtran}
\renewcommand\IEEEkeywordsname{Keywords}
\IEEEoverridecommandlockouts
\usepackage{cite}
\usepackage{pgfplots}
\usepackage{amsmath,amssymb,amsfonts}
\usepackage{algorithmic}
\usepackage{graphicx}
\usepackage{textcomp}
\usepackage{xcolor}
\usepackage{dblfloatfix}
\usepackage[justification=centering]{caption}
\usepackage{url}

\usepackage{breakurl}
\usepackage[breaklinks]{hyperref}
\def\BibTeX{{\rm B\kern-.05em{\sc i\kern-.025em b}\kern-.08em
    T\kern-.1667em\lower.7ex\hbox{E}\kern-.125emX}}
\begin{document}

\title{Human Activity Analysis and Recognition from Smartphones using Machine Learning Techniques}

\author{\IEEEauthorblockN{Jakaria Rabbi, Md. Tahmid Hasan Fuad, Md. Abdul Awal}
\IEEEauthorblockA{\textit{Department of Computer Science and Engineering} \\
\textit{Khulna University of Engineering \& Technology}\\
Khulna-9203, Bangladesh 
\\
jakaria\_rabbi@cse.kuet.ac.bd, mdtahmidhasanfuad@gmail.com, awal@cse.kuet.ac.bd}
}

\maketitle 

\begin{abstract}
Human Activity Recognition (HAR) is considered a valuable research topic in the last few decades. Different types of machine learning models are used for this purpose, and this is a part of analyzing human behavior through machines. It is not a trivial task to analyze the data from wearable sensors for complex and high dimensions. Nowadays, researchers mostly use smartphones or smart home sensors to capture these data. In our paper, we analyze these data using machine learning models to recognize human activities, which are now widely used for many purposes such as physical and mental health monitoring. We apply different machine learning models and compare performances. We use Logistic Regression (LR) as the benchmark model for its simplicity and excellent performance on a dataset, and to compare, we take Decision Tree (DT), Support Vector Machine (SVM), Random Forest (RF), and Artificial Neural Network (ANN). Additionally, we select the best set of parameters for each model by grid search. We use the HAR dataset from the UCI Machine Learning Repository as a standard dataset to train and test the models. Throughout the analysis, we can see that the Support Vector Machine performed (average accuracy 96.33\%) far better than the other methods. We also prove that the results are statistically significant by employing statistical significance test methods.
\end{abstract}


\section{Introduction}
Human beings can realize others' psychological state and personality by observing their daily activities. Following this pattern, the researchers want to predict human behavior using machines, and Human Behavior Recognition (HAR) comes as an active research topic. This has become one of the important research topics in machine learning and computer vision. Though motion data collection was hard in previous days, current technological developments help researchers capture the data as they can now use portable devices, including smartphones, music players, smartwatches, or smart home sensors. Especially, motion sensor embedded smartphones, like accelerometers, gyroscope, etc. bring a new era for activity recognition. Researchers use various machine learning and deep learning techniques including Naive Bayes (NB) \cite{bao2004activity}, Decision Tree \cite{tapia2004activity}, Support Vector Machine (SVM) \cite{zhu2012context}\cite{chathuramali2012faster}, Nearest Neighbor (NN) \cite{wu2012classification}, Hidden Markov Model (HMM) \cite{duong2005activity} Convolutional Neural Network (CNN) \cite{bevilacqua2018human} etc. to analyze the sensor data and recognize human activity. 

HAR is the problem of classifying day-to-day human activity using data collected from smartphone sensors. Data are continuously generated from the accelerometer and gyroscope, and these data are instrumental in predicting our activities such as walking or standing. There are lots of datasets and ongoing research on this topic. In \cite{lara2012survey}, the authors discuss wearable sensor data and related works of predictions with machine learning techniques. Wearable devices can predict an extensive range of activities using data from various sensors. Deep Learning models are also being used to predict various human activities \cite{plotz2018deep}. Nowadays, people use smartphones almost all the time and use many wearable devices. Through these devices, physical and mental health can be monitored by predicting human activity without specialized and costly medical equipment, and nowadays, it is an efficient, cheap, and safe way to do this as the COVID-19 (Coronavirus disease 2019) pandemic is ongoing.

In this paper, we have selected a dataset from the UCI machine learning repository \cite{MLrepo} to calculate the accuracy of five machine learning models and to perform some statistical significance tests. We use the Decision Tree (DT), Random Forest (RF), LR, SVM, and Artificial Neural Network (ANN) with a hidden layer to predict human activity from mobile data. In the dataset, the human activities are classified into six categories: walking, walking upstairs, walking downstairs, sitting, standing, and laying, and the data were collected from two different smartphone sensors, which are accelerometer and gyroscope. We ran experiments to see the performance of the models based on classification results and tried to get the highest possible accuracy from these models. Here, we tried to get the highest possible performance by these algorithms by parameter tuning, cross-validation, and finally, comparing the result with two statistical significance tests to get the winner algorithm. 
\par
The remainder of this paper is organized as follows. Section II
describes the related works, and section III discusses the dataset. Section IV discusses the machine learning models used in this paper, section V describes the experimental methods, and section VI presents the experimental results. Eventually, section VII concludes the paper.

\section{Related Works}
Since HAR becomes an important research topic, the researchers use different Machine Learning methods, such as Naive Bayes, Logistic Regression, Decision Trees, Support Vector Machines, Nearest Neighbor, Hidden Markov Model, CNN, RNN (Recurrent Neural Network), etc. for recognizing human activities. D. Singh et al. used RNN in their work for human activity recognition. They use smart home sensors to collect data and applied a Long Short Term Memory (LSTM) RNN. ANN is used for prototyping data acquisition module \cite{oniga2014human}. A Anjum et al. \cite{anjum2013activity} use smartphone sensors over detectable physical activities, including walking, running, climbing, cycling, driving, etc. They compare Naive Bayes, Decision Tree, KNN (K Nearest Neighbours), and SVM to calculate accuracy. They achieve a greater than 95\% true positive rate and less than 1.5\%  false positives rate.

Z Chen et al. \cite{chen2017robust} propose a robust HAR system based on coordinate transformation and PCA (CT-Principal Component Analysis) and online SVM. CT-PCA scheme is used to reduce the effect of orientation variations. Their presented OSVM is independent, which uses a tiny portion of data from the unseen placement. Sun et al. \cite{sun2010activity} propose a HAR approach with varying position and orientation through smartphone accelerometer data. They use their acceleration magnitude as the fourth data dimension. They also use generic SVM and location-specific SVM in the model.

Deep Learning approaches require a considerable amount of training data; therefore, many researchers mainly use generic machine learning approaches \cite{ML123}. Hence, in our paper, we focus on the generic approaches to analyze and detect human activity.

\section{Dataset}

\subsection{Data Exploration and Exploratory Visualization}
The dataset is taken from UCI machine learning repository\cite{MLrepo}. The dataset contains the information from 30 volunteers within the age range: 19-48. Each volunteer performs six activities: 
\begin{itemize}
    \item Walking
    \item Walking upstairs
    \item Walking downstairs
    \item Sitting
    \item Standing
    \item Laying
\end{itemize}
Hence, the dataset has six labels to predict. The dataset consists of 561 feature vectors with time and frequency domain variables. These features come from the following data:
\begin{itemize}
    \item Gravitational acceleration for x, y, and z axes 
    \item Body acceleration data files for x, y, and z axes
    \item Body gyroscope data files for x, y, and z axes
\end{itemize}
We also have the ID of an individual volunteer for each record. A total of 10299 records and training and test dataset splits are 70\% and 30\%. The 6 classes are converted to numerical values sequentially: [1, 2, 3, 4, 5, 6]. 

\begin{figure}[ht]
\centerline{\includegraphics[width=86mm,scale=1.0]{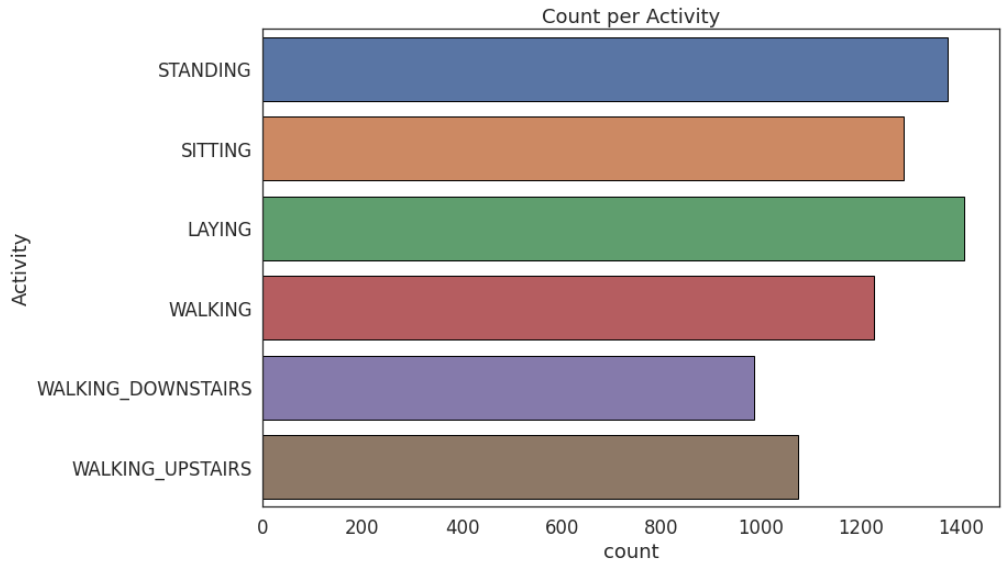}}
\caption{Different human activities with their count in the training dataset.}
\label{Figure}
\end{figure}
Figure 1 shows the total number of different human activities in training data, and figure 2 shows the distribution of two types of activity (static: sitting, standing, and laying – dynamic: walking, walking\_downstairs, and walking\_upstairs) using tBodyAccMag-mean() feature, which is taken from the accelerometer of smartphones. It is clear that the two types of activities are easily separable. There are no missing and duplicate values.
\begin{figure}[ht]
\centerline{\includegraphics[width=86mm,scale=1.0]{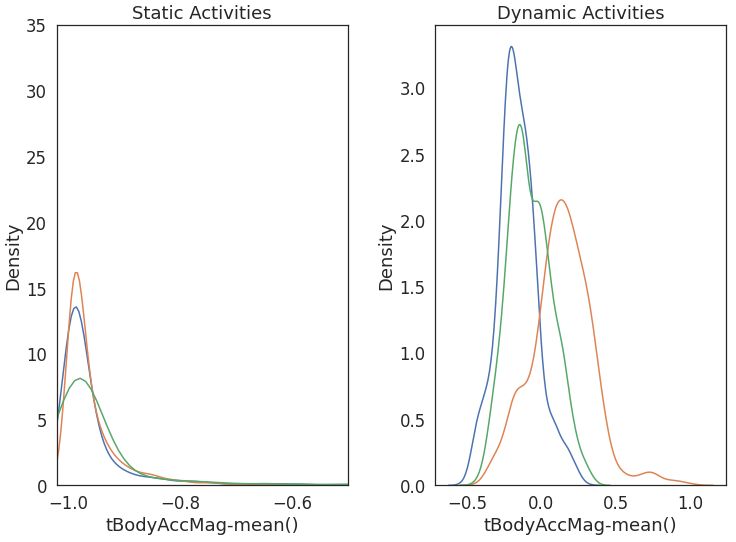}}
\caption{Static and dynamic activity pattern}
\label{Figure}
\end{figure}

\section{Machine Learning Models}
\subsection{Support Vector Machine}
Support Vector Machine (SVM) is used as one of the selected algorithms. SVM is a discriminative classifier determined by a separating hyperplane. If training data is fed to the SVM, it gives an optimal hyperplane that classifies new data. This hyperplane is simply a line dividing a plane into two parts in two-dimensional space wherein each class lay on either side \cite{patel_2017}. There are SVM variants, and one is Support Vector Classifier (SVC) for more than two classes. In this paper, we use SVC with ‘linear,’ ‘rbf’ \cite{wikipedia_2020} and ‘sigmoid’ kernel. SVM usually provides good accuracy when the number of features is large. In our dataset, the number of features is large; hence we selected this algorithm.

\begin{table}[h]
    \caption{Parameters for DT}
    \label{Table1}
    \centering
    \begin{tabular}{c c c}
\hline         \textbf{Parameters} & \textbf{Values} & \textbf{Description}\\ [0.5ex]
         \hline\hline
         Criterion & [Gini, Entropy] & Measure the quality of the \\
         & & data split.\\ 
         & & Gini - Gini  impurity and\\ & & Entropy - information gain.\\
         \hline
         Max Depth & [0, 2, 4, 6, 8] & The maximum depth of the tree.\\
         \hline
    \end{tabular}
\end{table}

\subsection{Decision Tree}
A Decision tree is a tree-like formation of decisions and possibilities. It differentiates instances by sorting from root to leaves. Here, we check impurity using Gini and gain information from entropy. These measures the quality of the split. We use various max depths (0 to 8), which is the maximum depth of the tree.

\begin{table}[h]
    \caption{Parameters for RF}
    \label{Table2}
    \centering
    \begin{tabular}{l l l}
\hline         \textbf{Parameters} & \textbf{Values} & \textbf{Description}\\ [0.5ex]
         \hline\hline
         Criterion & [Gini, Entropy] & Measure the quality \\
         & & of the data split\\ 
         & & Gini - Gini impurity \\
         & & and  Entropy -  \\
         & & information gain\\
         \hline
        N Estimators & [10, 30, 60, 90, 120, 150] & Number of trees in \\
        & & the generated forest\\
         \hline
         Max Depth & [0, 2, 4, 6, 8] & The maximum depth  \\
         & & of the tree\\
         \hline
    \end{tabular}
\end{table}

\begin{table}[h]
    \caption{Parameters for SVM}
    \label{Table2}
    \centering
    \begin{tabular}{l l l}
         \textbf{Parameters} & \textbf{Values} & \textbf{Description}\\ [0.5ex]
         \hline\hline
         Kernel & [Linear, RBF, Sigmoid] & The type of kernel  \\
         & & to be used in the \\
         & & model\\
         \hline
         C & [0.1, 0.5, 1, 2, 5, 10, 100] & Penalty parameter C of \\
         & & the error term.\\
         \hline
    \end{tabular}
\end{table}

\begin{table}[h]
    \caption{Parameters for ANN}
    \label{Table2}
    \centering
    \begin{tabular}{l l l}
        \textbf{Parameters} & \textbf{Values} & \textbf{Description}\\ [0.5ex]
         \hline\hline
         Hidden layer sizes & [(10,), (50,), (100,)] & Number of hidden nodes\\
         \hline
        Alpha & [1e-4, 1e-3, 1e-2] & L2 regularization \\
        & & parameter \\
         \hline
        Learning rate & [1e-3, 1e-2, 1e-1] & Step size\\
         \hline
        Beta1 & [0.1, 0.5, 0.9] & Adam specific parameter\\
         \hline
        Beta2 & [0.1, 0.5, 0.9] & Adam specific parameter\\
         \hline
    \end{tabular}
\end{table}

\begin{table}[hb]
    \caption{Parameters for LR}
    \label{Table2}
    \centering
    \begin{tabular}{l l l}
    \hline     \textbf{Parameters} & \textbf{Values} & \textbf{Description}\\ [0.5ex]
         \hline\hline
         Regularizer & [L1, L2] & Specifies the type \\
         & & of regularization \\
         \hline
         C & [0.1, 0.5, 1, 2, 5, 10, 100] & Penalty parameter C \\
         & & of the error term.\\
         \hline
    \end{tabular}
\end{table}

\subsection{Random Forest}
Random Forest (RF) model is an ensemble of some decision trees. It combines the learning methods and increases the overall result. Here criterion measures the quality of a split. We use Gini for impurity checks, and the information is gained from entropy. We use N estimators and max depth as the parameter. N estimators are the number of trees in the generated forest, and max depth is the tree's maximum depth.

\subsection{Artificial Neural Network}
An artificial Neural Network having a single hidden layer and the sigmoid activation function is used for our experiment. Adam, a stochastic gradient-based optimizer \cite{kingma2014adam}, is used for weight optimization. The maximum epoch is 200, and the dataset is shuffled between each epoch. The ANN is selected because it is currently used in many learning problems.

\subsection{Benchmark Model}
We used Logistic Regression (LR) as a benchmark model, and it focuses to maximize the probability of the data. This is a basic model without much complexity like neural networks. The performance of LR improves if the data lies on the correct side of the separating hyperplane. We used the kernel trick and ‘lbfgs’ optimizer to compare the performance. Some previous research has some good results with that combination of LR \cite{cawley2008efficient}. Hence, we used this algorithm because LR, and its variants are the right choice for benchmarking.

\section{Experimental Methods}
\subsection{Data preprocessing}
Dataset has a train and test portion. There are no duplicate and missing values. Train and Test data are stored in the data frame initially. We tried to remove some features to reduce the feature space. We tried to select features by computing ANOVA F-value \cite{wikipedia_2021} for the dataset. We selected top 100, 200, 400, and 500 features with top ANOVA F-values.
\begin{figure}
\centering
\begin{tikzpicture}[scale=0.8]
\begin{axis}[
    xlabel={Number of Features},
    ylabel={Accuracy (\%)},
    xmin=0, xmax=600,
    ymin=0, ymax=100,
    xtick={100,300,200,400,500,600},
    ytick={0,20,40,60,80,100},
    legend pos=south east,
    ymajorgrids=true,
    xmajorgrids=true,
    grid style=dashed,
    width=9cm,
    height=5cm,
    every axis plot/.append style={thick}
]

\addplot[
    color=green,
    mark=square,
    ]
    coordinates {
    (100,83.3)(200, 85.7)(300,87.9)(400,90.6)(500,91.2)(561,93.6)};

\addplot[
    color=blue,
    mark=square,
    ]
    coordinates {
    (100,79.9)(200, 81.2)(300,82.7)(400,85.5)(500,87.1)(561,88.9)};

\addplot[
    color=red,
    mark=square,
    ]
    coordinates {
    (100,85.1)(200, 87.5)(300,88.3)(400,91.5)(500,92.1)(561,94.9)};

\addplot[
    color=orange,
    mark=square,
    ]
    coordinates {
    (100,81.8)(200, 85.7)(300,85.9)(400,87.3)(500,89.5)(561,92.8)};

\addplot[
    color=black,
    mark=square,
    ]
    coordinates {
    (100,65.7)(200, 67.6)(300,69.5)(400,72.5)(500,75.7)(561,78.9)};    

    \legend{LR,RF,SVM,ANN,DT}
    
\end{axis}
\end{tikzpicture}
\caption{Increasing accuracy of the five algorithms by increasing features.}
\label{Figure}
\end{figure}
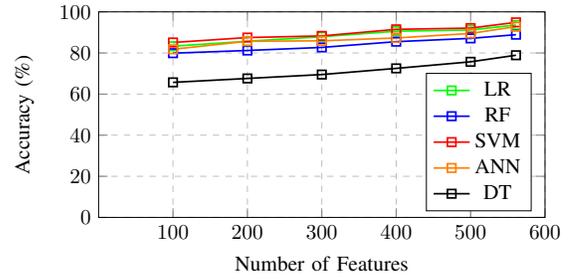
Nevertheless, each time we increased the features, the accuracy increased significantly for all five algorithms and the trend is depicted in figure 3. Therefore, we use all the features for training and testing. 
The learning is divided into two parts. Initially, the best parameters for all the algorithms are learned. Then, in the second step, the algorithms are compared. The dataset is evaluated by a K-fold cross-validation set, and we used k=5 for cross-validation. The data splitting is done by stratified sampling. We use 70\% training data for cross-validation purposes without touching the test data.
\begin{table}[t]
    \caption{Best parameters for the selected models}
    \label{Table2}
    \centering
    \begin{tabular}{c c}
\hline         \textbf{Models} & \textbf{Parameters} \\ [0.5ex]
         \hline\hline
         DT & Criterion: Gini, Max Depth: 4 \\
         \hline
         RF & Criterion: Gini, N Estimators: 90, Max Depth: 6\\
         \hline
         SVM & C: 1.0, Kernel: Linear\\
         \hline
         ANN & Alpha: 0.01, Beta1: 0.9, Beta2: 0.9,\\& Hidden layer sizes: (50,), Learning rate: 0.001\\
         \hline
         LR & C: 2, Regularizer: L1\\
         \hline
    \end{tabular}
\end{table}

\begin{table}[t]
    \label{Table4}
    \caption{Accuracies for all algorithms}
    \centering
    \begin{tabular}{c c}
     \hline
     \textbf{Model} & \textbf{Average Accuracy} \\[0.5ex]
     \hline\hline
     DT & 80.3\% \\
     \hline
     RF & 90.05\% \\
     \hline
     SVM & 96.33\% \\
     \hline
     ANN & 93.8\% \\
     \hline
     LR & 96.01\% \\
     \hline
    \end{tabular}
\end{table}
\subsection{Implementation}
A cross-validation dataset is used for an exhaustive grid search to find the best hyperparameters. The models learn from the dataset for all combinations of the listed parameters. Table I (for DT), Table II (for RF), Table III (for SVM), Table IV (for ANN), and Table V (for LR) have the parameter list that we use to get the best combination of parameters. The parameters are chosen from a range of values that work well in many problems. Accuracy is used as the scoring method. We use the python Scikit-learn library for the implementation. A detailed explanation of all parameters can be found in \cite{DT-SL}\cite{RF-SL}\cite{SVC-SL}\cite{ANN-SL}\cite{LR-SL}.

\subsection{Refinement}
After finding the best parameters, we evaluated the five algorithms on test data. Test data is divided into 10 random sets, and every set consists of 50\% of the data. Hence, we have 10 runs for every algorithm and calculated the mean and variance for these runs. The learning process to find the best parameters used only training data. Hence, we got an unbiased estimation for cross algorithm comparisons.
\par
The best parameters for the five algorithms are found after an exhaustive grid search.
All combination of the given parameters is used to find the best parameters After 5-fold cross-validation (stratified), we got the best parameters in table VI.

\begin{table}[h]
    \label{Table5}
    \caption{Average scores per class for DT}
    \centering
    \begin{tabular}{c c c c}
     \hline
     \textbf{Class} & \textbf{Precision} & \textbf{Recall} & \textbf{F1-score}  \\[0.5ex]
     \hline\hline
     1 & 0.76 & 0.86 & 0.81\\
     \hline
     2 & 0.86 & 0.79 & 0.84\\
     \hline
     3 & 0.93 & 0.88 & 0.91\\
     \hline
     4 & 0.85 & 0.88 & 0.87\\
     \hline
     5 & 0.87 & 0.83 & 0.85\\
     \hline
     6 & 0.83 & 0.84 & 0.84\\
     \hline
    \end{tabular}
\end{table}

\begin{table}[h]
    \label{Table6}
    \caption{Average scores per class for RF}
    \centering
    \begin{tabular}{c c c c}
     \hline
     \textbf{Class} & \textbf{Precision} & \textbf{Recall} & \textbf{F1-score}  \\[0.5ex]
     \hline\hline
     1 & 0.83 & 0.96 & 0.89\\
     \hline
     2 & 0.93 & 0.84 & 0.88\\
     \hline
     3 & 0.98 & 0.96 & 0.97\\
     \hline
     4 & 0.92 & 0.97 & 0.94\\
     \hline
     5 & 0.94 & 0.91 & 0.93\\
     \hline
     6 & 0.91 & 0.92 & 0.92\\
     \hline
    \end{tabular}
\end{table}
\begin{table}[h]
    \label{Table7}
    \caption{Average scores per class for ANN}
    \centering
    \begin{tabular}{c c c c}
     \hline
     \textbf{Class} & \textbf{Precision} & \textbf{Recall} & \textbf{F1-score}  \\[0.5ex]
     \hline\hline
     1 & 0.86 & 0.97 & 0.91\\
     \hline
     2 & 0.96 & 0.86 & 0.91\\
     \hline
     3 & 1.00 & 0.97 & 0.99\\
     \hline
     4 & 0.92 & 0.99 & 0.95\\
     \hline
     5 & 0.98 & 0.94 & 0.96\\
     \hline
     6 & 0.95 & 0.95 & 0.93\\
     \hline
    \end{tabular}
\end{table}

\subsection{Refinement with Statistical Significance Test}
We used hypothesis testing for this paper. A hypothesis is examined by estimating and inspecting a random sample of the population selected to be analyzed. We examined specific hypotheses using a random population pattern: one is the null and another one is the alternative hypothesis. The null hypothesis related to the equality between population parameters, and as an example, a null hypothesis assumes that the population mean return is the same to zero. Alternatively, the alternative hypothesis is efficaciously the reverse of a null hypothesis, and as an instance, the population mean return is not identical to zero. Therefore, they can not happen simultaneously, and only one hypothesis can be correct. We tested all hypotheses using the following four-step process:

\begin{itemize}
    \item At first, the we have to state the two hypotheses so that only one can be accurate. 
    \item Then, there is a plan to outline how the data are determined.  
    \item After that, the plan is carried out, and the sample data are analyzed.

\begin{table}[h]
    \label{Table8}
    \caption{Average scores per class for SVM}
    \centering
    \begin{tabular}{c c c c}
     \hline
     \textbf{Class} & \textbf{Precision} & \textbf{Recall} & \textbf{F1-score}  \\[0.5ex]
     \hline\hline
     1 & 0.90 & 0.97 & 0.94\\
     \hline
     2 & 0.96 & 0.89 & 0.92\\
     \hline
     3 & 1.00 & 1.00 & 1.00\\
     \hline
     4 & 0.95 & 1.00 & 0.97\\
     \hline
     5 & 0.99 & 0.98 & 0.98\\
     \hline
     6 & 0.98 & 0.95 & 0.97\\
     \hline
    \end{tabular}
\end{table}
\begin{table}[h]
    \label{Table9}
    \caption{Average scores per class for LR}
    \centering
    \begin{tabular}{c c c c}
     \hline
     \textbf{Class} & \textbf{Precision} & \textbf{Recall} & \textbf{F1-score}  \\[0.5ex]
     \hline\hline
     1 & 0.89 & 0.98 & 0.93\\
     \hline
     2 & 0.98 & 0.86 & 0.91\\
     \hline
     3 & 1.00 & 1.00 & 1.00\\
     \hline
     4 & 0.95 & 1.00 & 0.97\\
     \hline
     5 & 1.00 & 0.97 & 0.99\\
     \hline
     6 & 0.97 & 0.95 & 0.96\\
     \hline
    \end{tabular}
\end{table}
 
\begin{table*}[ht]
    \label{Table10}
    \caption{Welch’s t-test values for the top three models}
    \centering
    \begin{tabular}{c c c c c c c} 
     \hline
     \textbf{Models} & SVM(t-value) & SVM(p-value) & ANN(t-value) & ANN(p-value) & LR(t-value) & LR(p-value) \\ [0.5ex]
     \hline\hline
     SVM (t-value $|$ p-value) & 0 & 1 & 7.44 & 6.71e-07 & 2.31 & .0324\\ 
     \hline
     ANN (t-value $|$ p-value) & -7.44 & 6.71e-07 & 0 & 1 & -5.88 & 1.42e-05 \\
     \hline
     LR (t-value $|$ p-value) & -2.31 & .0324 & 5.88 & 1.42e-05 & 0 & 1 \\
     \hline\\
    \end{tabular}
\end{table*}

\begin{table*}[ht]
    \label{Table11}
    \caption{5 by 2-fold Cross Validation t paired test}
    \centering
    \begin{tabular}{c c c c c c c} 
     \hline
     Models & SVM(t-value) & SVM(p-value) & ANN(t-value) & ANN(p-value) & LR(t-value) & LR(p-value) \\ [0.5ex]
     \hline\hline
     SVM (t-value $|$ p-value) & 0 & 1 & 2.72 & 4.1e-02 & 3.38 & .0195\\ 
     \hline
     ANN (t-value $|$ p-value) & -4.65 & 5.5e-03 & 0 &	1 &	-6.11 &	1.7e-03 \\
     \hline
     LR (t-value $|$ p-value) & -3.81 & .0124 & 2.73 & 4.0e-02 & 0 & 1 \\
     \hline\\
    \end{tabular}
\end{table*}

   \item Finally, the result is analyzed and take a decision whether to reject the null hypothesis, or state that the null hypothesis can be possible with the given data \cite{majaski_2021}.
\end{itemize}

The algorithms are compared using Welch’s t-test \cite{wikipedia_2021_1} method because the algorithms' variances are different. We can find output from the t-test that shows whether the algorithms have significant changes in their performance measurements. The t-test show the importance of the differences between groups. It can explain whether the differences measured in means, could have happened by chance. The two-tailed t-test is done pairwise for every algorithm with $H_{0}: \mu_{0} = \mu_{1}$
and significance level $\alpha = 0.05$ \cite{wikipedia_2020_1}
to produce a ranking between the algorithms and for finding the winner algorithm. Here, H\textsubscript{0} is the null hypothesis, and µ\textsubscript{0} and µ\textsubscript{1} are the means of two groups of population. Also, the algorithms are compared using 5 times 2-fold cross-validated paired t-test \cite{raschka} as suggested by the researchers in \cite{dietterich1998approximate}. 

\section{Experimental Results}
We used 5-fold stratified cross-validation and grid search to find the best hyperparameters. Then run the models with the best parameters on test data and finally run the statistical significance test.

\subsection{Evaluation Metrics}
Accuracy is calculated by dividing the number of correct predictions by the total number of predictions. This metric is fundamental as we want to know whether a human is walking or sitting correctly. Accuracy is the primary metric to distinguish between our models. In equation 1, we get an accuracy score using True Positive (TP), True Negative (TN), False Positive (FP), and False Negative (FN). 
\begin{equation}
Accuracy = \frac{TP+TN}{TP+TN+FP+FN} 
\end{equation}
Additionally, we use Precision, Recall, and F1 score as the evaluation metrics. The equation of these three metrics is given in equations 2, 3, and 4.
\begin{equation}
Precision = \frac{TP}{TP+FP} 
\end{equation}
\begin{equation}
Recall = \frac{TP}{TP+FN} 
\end{equation}
\begin{equation}
F1 = \frac{2*Precision*Recall}{Precision+Recall}
\end{equation}

\subsection{Model Evaluation \& Validation}
Using the derived best parameters, the best model for each algorithm is formulated. Then the models are evaluated on the test data. Test data is divided into 5 random folds (stratified), and each fold contains the whole dataset. We got the average accuracy for the five different models using the data.

In Table VII, we can see the accuracies for the algorithms. Decision Tree and Random forest have low accuracy compared to the other five algorithms. ANN performs well but has less accuracies compare to the other two algorithms. We can see the average precision, recall, and F1 score per class from tables VIII, IX, X, XI, and XII. There are 6 classes, and all scores per class are in the tables. We also understand that SVM and LR perform better than DT, RF, and ANN by inspecting the results from the tables mentioned above. We also tried multiple hidden layers with more hidden nodes, but the accuracy dropped and the phenomenon is depicted in figure 4.

\begin{figure}
\centering
\begin{tikzpicture}[scale=0.8]
\begin{axis}[
    xlabel={Number of Hidden Layers},
    ylabel={Accuracy (\%)},
    xmin=0, xmax=5,
    ymin=0, ymax=100,
    xtick={1,2,3,4,5},
    ytick={0,20,40,60,80,100},
    legend pos=south east,
    ymajorgrids=true,
    xmajorgrids=true,
    grid style=dashed,
    width=9cm,
    height=3.5cm,
    every axis plot/.append style={thick}
]

\addplot[
    color=blue,
    mark=square,
    ]
    coordinates {
    (1,93.8)(2, 90.7)(3,89.1)(4,85.6)(5,80.1)};
    
     \legend{ANN}
    
\end{axis}
\end{tikzpicture}
\caption{Decreasing accuracy of ANN by increasing the number of hidden layers.}
\label{Figure}
\end{figure}
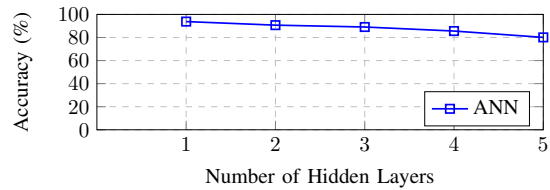

Hence, we did not include that model for comparison. We used the kernel trick implementation for logistic Regression, but the accuracy is not significantly sufficient, and it takes more time than simple LR. Therefore, for simplicity, we did not include kernels in the LR. Both SVM and LR have closer accuracy and precision-recall scores per class. Nevertheless, we cannot decide the best model without a statistical significance test. 

\subsection{Statistics for Justification}
Test data is divided into 10 random folds (stratified), and each fold contains 50\% of the data. Then we run the top three best models with the best parameters. After getting the result from 10 runs for each of the three models, we ran a two-tailed Welch’s t-test on the results from the top three models. We got the result for the three models and shown in table XIII. From Table XIII, we understand that SVM is the winner algorithm. Our hypothesis 
$H_{0}: \mu_{0} = \mu_{1}$ has $\alpha = 0.05$
and from this table we see that every p-value is less than alpha. Therefore, we can reject the hypothesis. So, we can conclude that the results are significantly different. Hence, SVM is the best algorithm here.
We also tested with a 5 by 2-fold cross-validation t paired test to confirm the winner algorithm. In several research papers \cite{paired_t_test}\cite{paired_t_test_1}\cite{paired_t_test_2}, we found this as a recommended test. In this test, the whole dataset (training and test data) is divided into two random folds and run 5 times. Then the results are collected, and the t statistic is calculated. In table XIV, the statistic is shown. It is clearly visible that the hypothesis is rejected every time, and the results are significantly different according to the test.
Table 9 shows that the t-value $|$ p-value pair is different for the same pair if the order is different. Because 5 times 2-fold cross-validation makes the outcome different. We can conclude that the SVM algorithm is the winner for this dataset.

\section{Conclusion}
In conclusion, it is evident that SVM performs better than the other two algorithms, though LR is the closer in case of accuracy. ANN cannot compete with SVM for this dataset. In most of the classification problems, ANN and deep ANN performs well. For this dataset, deep neural network architectures or convolutional neural network might increase the performance. We think we will explore the path in the near future. We also want to explore the performance of other machine learning algorithms in the future. In this experiment, we understand that simple machine learning algorithms can do well with proper parameter tuning, and we can determine the significance of the result by the statistical testing.

\bibliographystyle{IEEEtran}
\bibliography{Human_Activity_Analysis.bbl}

\vspace{12pt}

\end{document}